\documentclass[11pt,a4paper]{article}
\usepackage[hyperref]{ranlp2023}
\usepackage{times}
\usepackage{latexsym}

\usepackage{graphicx}
\usepackage{booktabs}

\usepackage{microtype}

\aclfinalcopy 
\def\aclpaperid{***} 

\title{Detecting Text Formality: A Study of Text Classification Approaches}

\author{
\textbf{Daryna Dementieva\textsuperscript{1}}, \textbf{Nikolay Babakov}\textsuperscript{2}, \textbf{and}
\textbf{Alexander Panchenko\textsuperscript{3,4}} \\
\textsuperscript{1}Technical University of Munich \\ \textsuperscript{2}Centro Singular de Investigación en Tecnoloxías Intelixentes (CiTIUS), \\ Universidade de Santiago de Compostela \\
\textsuperscript{3}Skolkovo Institute of Science and Technology, \textsuperscript{4}Artificial Intelligence Research Institute\\ 
\href{mailto:daryna.dementieva@tum.de}{\texttt{\small daryna.dementieva@tum.de}},
\href{mailto:nikolay.babakov@usc.es}{\texttt{\small nikolay.babakov@usc.es}},
\href{mailto:a.panchenko@skol.tech}{\texttt{\small a.panchenko@skol.tech}}
}

\date{}

\begin{document}
\maketitle
\begin{abstract}
Formality is one of the important characteristics of text documents. The automatic detection of the formality level of a text is potentially beneficial for various natural language processing tasks.
Before, two large-scale datasets were introduced for multiple languages featuring formality annotation---GYAFC and X-FORMAL. However, they were primarily used for the training of style transfer models. At the same time, the detection of text formality on its own may also be a useful application. This work proposes the first to our knowledge systematic study of formality detection methods based on statistical, neural-based, and Transformer-based machine learning methods and delivers the best-performing models for public usage. We conducted three types of experiments -- monolingual, multilingual, and cross-lingual. The study shows the overcome of Char BiLSTM model over Transformer-based ones for the monolingual and multilingual formality classification task, while Transformer-based classifiers are more stable to cross-lingual knowledge transfer. 

\end{abstract}

\section{Introduction}
\label{sec:intro}

According to \newcite{joos1976five}, five different types of text formality are commonly identified in Linguistics: frozen style, formal style, consultative style, casual style, and intimate style. The correct use of style is important for fluent human communication and, therefore, for fluent human-to-machine communication and various Natural Language Processing (NLP) systems. 

The examples of formal and informal samples for English, Brazilian Portuguese, French, and Italian languages are provided in Table~\ref{tab:examples}. As we can see, for informal sentences, several attributes are typical -- the usage of spoken abbreviations (for instance, \textit{lol}), non-standard capitalization of words (all words are written in upper case), and lack of punctuation. On the contrary, in formal samples, all necessary punctuation is present, standard capitalization is used, some opening expressions can be observed in sentences (for example, \textit{in my opinion}).

These examples are taken from two only currently available text collections with formality annotation are GYAFC~\cite{rao-tetreault-2018-dear} and X-FORMAL~\cite{briakou-etal-2021-ola}. However, these datasets were primarily introduced for the task of style transfer. In this paper, we propose to look at these data sets from a different angle. 
Even for the evaluation of the results of formality style transfer, we need to calculate \textit{style transfer accuracy}. While there is ongoing work of developing automatic evaluation metrics for formality style transfer in general \cite{lai-etal-2022-human}, this work introduces a systematic evaluation of formality style classifiers.







\begin{table*}[ht!]
    \centering
    \footnotesize
    \begin{tabular}{p{1.5cm}p{13.5cm}}
        \hline
        \multicolumn{2}{c}{\textbf{English}} \\
        \hline
        Formal & \textit{I enjoy watching my companion attempt to role-play with them.}\\
        Informal & \textit{lol i love watchin my lil guy try to act out the things wiht them}\\ 
        \hline
        \multicolumn{2}{c}{\textbf{Brazilian Portuguese}} \\
        \hline
        Formal & \textit{Na minha opinião, Beyonce, porque ela é mais jovem e uma dançarina melhor.}\\
        & \textcolor{gray}{\textit{In my opinion, Beyonce, because she's younger and a better dancer. }} \\
        Informal & \textit{BEYONCE PORQUE ELA É MAIS JOVEM E PODE DANÇAR MELHOR}\\ 
        & \textcolor{gray}{\textit{BEYONCE BECAUSE SHE IS YOUNGER AND CAN DANCE BETTER}} \\
        \hline
        \multicolumn{2}{c}{\textbf{French}} \\
        \hline
        Formal & \textit{Bien sûr, c'est Oprah, parce qu'elle fournit de meilleurs conseils depuis plus longtemps.}\\
        & \textcolor{gray}{\textit{Of course, it's Oprah, because she's been providing better advice for longer. }} \\
        Informal & \textit{oprah bien sûr parce qu'elle donne de meilleurs conseils et l'a fait plus longtemps}\\ 
        & \textcolor{gray}{\textit{oprah of course because she gives better advice and did it longer }} \\
        \hline
        \multicolumn{2}{c}{\textbf{Italian}} \\
        \hline
        Formal & \textit{King ha una canzone su questo, si chiama ``Solo tua madre ti ama".}\\
        & \textcolor{gray}{\textit{King has a song about this, it's called ``Only Your Mother Loves You."}} \\
        Informal & \textit{King aveva una canzone su questo - Solo la tua Madre ti ama (e vedere potrebbe essere anche jiving).}\\ 
        & \textcolor{gray}{\textit{King had a song about this - Only your Mother loves you (and seeing could be jiving too).}} \\
        \hline
    \end{tabular}
    \caption{Examples of samples from GYAFC and X-FORMAL datasets for four languages: English, Brazilian Portuguese, French, and Italian.}
    \label{tab:examples}
\end{table*}

In this paper, we aim at closing the gap by proposing a comprehensive computational study of various text categorization approaches. Namely, we argue that NLP practitioners will be benefiting from the knowledge of answers to the following questions:

\begin{description}
    \item[\textbf{Q1}:] What is the state-of-the-art for monolingual English formality classification?

    \item[\textbf{Q2}:] Can we train multilingual model for simultaneous formality detection on several languages?

    \item[\textbf{Q3}:] To what extent is cross-lingual transfer between pre-trained classifiers possible (if the phenomenon of formality is expressed similarly in various languages)?
\end{description}

To answer these questions, we present monolingual, multilingual, and cross-lingual experiments for formality classification for four languages---English, Brazilian Portuguese, French, and Italian.\footnote{\href{https://huggingface.co/s-nlp/mdeberta-base-formality-ranker}{https://huggingface.co/s-nlp/mdeberta-base-formality-ranker. Accessed 15 July 2023}}


    
    
    

\section{Related Work}


\subsection{Formality Datasets}

Formality detection was first investigated by \newcite{pavlick2016empirical} where the authors created datasets of formal and informal sentences sourced from news, emails, blogs, and community answering services. The sentences were scored by a formality rating. 

In \cite{rao-tetreault-2018-dear}, a dataset called GYAFC for formality style transfer evaluation has been proposed for the English language. 
%
%
After that, in \cite{briakou-etal-2021-ola}, the authors proposed the first multilingual dataset containing formality annotation, called X-FORMAL. The dataset features Brazilian Portuguese, Italian, and French languages and is structurally similar to the English GYAFC. 


While the original papers on GYAFC and X-FORMAL provided extensive experimental results with these datasets, they all were focused on the style transfer setting and did not study the formality detection task. Our study instead focuses on text classification using these datasets. 

\subsection{Text Classification}

Text categorization is well-established NLP task with dozens of applications ranging from topic categorization to fake news detection, with the first works dating back to the late 80-s ~\cite{hayes1988news,lewis1991evaluating}. 

\newcite{sebastiani2002machine} provides a comprehensive survey on the ``classic'' methods on text categorization. Much more specialized text categorization methods have been developed so far, notably neural models such as CharCNN~\cite{DBLP:conf/nips/ZhangZL15} or more advanced solutions based on large pre-trained transformer networks, such as BERT~\cite{sun2019fine}. In \cite{DBLP:conf/iri/LiWBHGMS22}, Formality-LSTM and Formality-BERT were proposed to detect formality in answers, blogs, emails, and news. 

To overcome the privilege of only monolingual models development, several multilingual pre-trained language models were introduced. In our experiments, we adjusted for sequence classification task mT5 \cite{xue2021mt5} (covers 101 languages) and mBART \cite{tang2020multilingual} (covers 50 languages) models. 

\section{Datasets}
Here, we provide the detailed description of the data---nature of the texts and general datasets' statistics---used for the experiments.


\begin{table*}

\centering
\footnotesize
\begin{tabular}{l|c|c c|c c}
\hline
& & \multicolumn{2}{|c|}{Informal to Formal} & \multicolumn{2}{|c}{Formal to Informal} \\\hline
\textbf{} & \textbf{Train} & \textbf{Tune} & \textbf{Test} & \textbf{Tune}& \textbf{Test} \\ \hline

Entertainment and Music domains & 105\,190  &  2\,877  &  1\,416  & 2\,356 &  1\,082  \\
Family and Relationships domains & 103\,934 &  2\,788  & 1\,332  & 2\,247 &  1\,019  \\
All domains, no duplicates & 204\,365 & 29\,132 & 10\,710 & 19\,448 & 9\,031 \\

\hline
\end{tabular}
\caption{Statistics of the GYAFC dataset.}
\label{tab:gyafc_stats}
\end{table*}

\begin{table*}
\centering
\footnotesize
\begin{tabular}{l|c|c|c|c}
\hline

\textbf{Dataset} & \textbf{Language} & \textbf{\# texts} & \textbf{\# formal texts} & \textbf{\# informal texts} \\ \hline

GYAFC~\cite{rao-tetreault-2018-dear} & EN &  204\,365 & 102\,182 & 102\,183 \\

X-FORMAL~\cite{briakou-etal-2021-ola} & FR+IT+BR &  338\,763 & 168\,099 & 170\,664 \\

X-FORMAL~\cite{briakou-etal-2021-ola} & FR &  112\,921 & 56\,033 & 56\,888 \\

X-FORMAL~\cite{briakou-etal-2021-ola} & IT &  112\,921 & 56\,033 & 56\,888 \\

X-FORMAL~\cite{briakou-etal-2021-ola} & BR &  112\,921 & 56\,033 & 56\,888 \\

\hline
\end{tabular}
\caption{Statistics of the GYAFC ans X-FORMAL datasets.}
\label{tab:xformal_stats}
\end{table*}

\subsection{English: GYAFC}

GYAFC---English dataset---contains 104\,365 pairs of formal and informal texts obtained from Yahoo Answers. It consists of two parts split between Entertainment \& Music and Family \& Relationship categories. Firstly, informal texts were collected. Then, they were manually rewritten to create a formal alternative in the parallel pairs. The dataset also contains the tune and test text pairs. The creation of these pairs involved stricter control over the quality of translation. These pairs were also split in half between informal to formal translations and formal to informal translations. 

Descriptive statistics of both parts of the dataset are presented in Table~\ref{tab:gyafc_stats}. In our experiments, we use the dataset corresponding to the ``All domains, no duplicates''. 

\subsection{French, Italian, and Brazilian: X-FORMAL}

The X-FORMAL dataset \cite{briakou-etal-2021-ola} was created on the basis of the GYAFC dataset described in the section above. The goal of this dataset is to cover formality in multiple languages. More specifically, there are three languages included: Brazilian Portuguese (BR), French (FR), and Italian (IT). All these parts of the X-FORMAL dataset were created by translating the original GYAFC dataset from English to target languages. The dataset consists of 338\,763 samples in four languages. More detailed statistics of the X-FORMAL dataset are presented in Table \ref{tab:xformal_stats}.

In both datasets, the mean amount of tokens in samples is $10 \pm 4$ meaning that in the majority of cases we work with one-sentence samples.

\section{Text Classification Models}
\label{sec:methods}

Following \cite{lai-etal-2022-human}, we address the formality detection as text classification task. We experiment with several state-of-the-art models optimizing their hyper-parameters. A detailed description of these most successful models is presented below.

\subsection{Linguistic-Based Baselines} 

Firstly, we build with a heuristic approach based on punctuation presence in the text and capitalization of the first word denoted as ``punctuation + capitalization''. It is natural to expect that all sentences in formal style should start with a capital letter and end with the presence of some punctuation. For informal sentences, that can be missed.

Secondly, we test the classic bag-of-word representation used commonly in various text categorization tasks. In addition, we also tested another simple and common word vector representation: a mean of dense vector representations. For this variant, for the embeddings, we use pre-trained fastText vectors~\cite{bojanowski-etal-2017-enriching} for both English and multilingual experiments.\footnote{\href{https://fasttext.cc}{https://fasttext.cc. Accessed 10 January 2023}} 

On top of these types of features, we use Logistic Regression (LR), a linear model that is a workhorse for many text classification tasks. 

\subsection{Models based on Convolutional Neural Networks (CNNs)}

To get another way of vector representations for texts, we utilize Universal Sentence Encoder \cite{DBLP:journals/corr/abs-1907-04307}. This encoder is trained on 16 languages and is competitive with state of the art on semantic retrieval, translation pair bitext retrieval, and retrieval question answering tasks. Then, the obtained vectors is fed into a CNN model that consists of 2 CNN layers. The encoder is trained using Multi-task Dual Encoder Training similar to \cite{cer-etal-2018-universal}, and \cite{chidambaram2019learning} with a single encoder supporting multiple downstream tasks.



\subsection{BiLSTMs}

We also experiment with RNN for text classification as they have shown superior results in many tasks, with bidirectional LSTMs being the most popular choice.~\cite{DBLP:journals/access/HameedG20,isnain2020bidirectional,DBLP:journals/corr/abs-1811-02906} More specifically, we test two input representations for RNNs: character-based and token/word-based. \textit{Char BiLSTM} consists of an Embedding layer on chars followed with bidirectional LSTM layers \cite{graves2005framewise}. We tune several model configurations: embeddings size, number of BiLSTM layers, BiLSTM hidden layer size. According to our experiments, we achieved the best result with an embeddings size of $50$, the number of BiLSTM layers of $2$, and BiLSTM hidden layer size of $50$. 

In the \textit{Word BiLSTM}, the embedding layer is replaced by a pretrained fastText embedding layer, and \textsf{\small wordpunct\_tokenize} from NLTK is used to tokenize the text. We tune the same configurations as the Char BiLSTM and used Fastext 300d embeddings. According to our experiments, the best results were achieved with Fastext uncased 100d, the number of BiLSTM layers of $1$, and the BiLSTM hidden size of $50$.

\subsection{ELMo}

In addition to the BiLSTM architecture described above where pre-trained word embeddings are used, we also test the popular architecture for obtaining contextualized vector representations of tokens called ELMo~\cite{peters-etal-2018-deep}. It consists of two BiLSTM layers trained on character representations of the input text. 

We use a BiLSTM layer on top of the sequence of token embeddings obtained from ELMo, followed by two Dense layers and two Dropout layers. 

\subsection{Transformer-based Models}

More recently, the state-of-the-art in a variety of text classification tasks was achieved by models based on the deep neural networks based on the Transformer blocks~\cite{DBLP:conf/nips/VaswaniSPUJGKP17} pre-trained on a large text corpora.
In our work, we experiment with several such state-of-the-art models listed below. 

\paragraph{BERT}  We utilize BERT~\cite{devlin2019bert} and its distilled version---DistilBERT \cite{DBLP:journals/corr/abs-1910-01108}---models for monolingual English formality classification. We use \textsf{\small base} \textsf{\small uncaused} and \textsf{\small cased} versions of the mentioned models to check the contribution of the letter capitalization. Also, we test the next generations of BERT-like models---RoBERTa \textsf{\small roberta-base} \cite{DBLP:journals/corr/abs-1907-11692} and Deberta \textsf{\small deberta-base/large} \cite{DBLP:conf/iclr/HeLGC21}.


\paragraph{XLNet} This model integrates ideas of autoregressive language models \cite{DBLP:conf/nips/YangDYCSL19}. The usage of all possible permutations of the factorization order allows to use of bidirectional contexts of each token and outperforms the BERT model on several tasks. We fine-tune \textsf{\small xlnet-base-cased} version of this type of model.

\paragraph{GPT2}

In contrast to the mentioned above models, which all rely on the encoder of the original transformer architecture~\cite{DBLP:conf/nips/VaswaniSPUJGKP17} the GPT2 model~\cite{radford2019language} is based on the decoder of the Transformer. We utilize the raw hidden states from the last transformer block of the model \textsf{\small gpt2} to feed it into a linear classification head.

\paragraph{Multilingual Language Models}

Experiments on the multilingual X-FORMAL dataset require additional multilingual word embeddings extraction and text classification models. For this purpose, we use multilingual available analogues of afore mentioned models where all needed languages are supported. Firstly, we use mBERT~\cite{devlin2018bert} (and its distilled version of it as well---mDistilBERT) and mDeBERTa that was pretrained on 104 languages with the largest Wikipedia corpus (\textsf{\small bert/distilbert-base-multilingual-cased} and \textsf{\small mdeberta-v3-base} versions). Then, we experiment with multilingual version of XLNet---XLM-R \cite{conneau-etal-2020-unsupervised} (\textsf{\small xlm-roberta-base}, 100 languages). In addition, we provide the results of multilingual encoder-decoder-based models---mT5 \cite{xue2021mt5} (\textsf{\small mt5-base}, 101 languages) and mBART \cite{tang2020multilingual} (\textsf{\small mbart-large-50}, 50 languages).

\begingroup
\renewcommand{\arraystretch}{1.2}
\begin{table*}[h!]
    \centering
    \footnotesize
    \begin{tabular}{l|c|c|c|c|c|c|c}
        \toprule
         & & \multicolumn{3}{|c|}{\textbf{Formal}} & \multicolumn{3}{|c}{\textbf{Informal}} \\
        \hline
        \textbf{Text Representation Model}  & Accuracy & Precision & Recall & F1 & Precision & Recall & F1 \\
        \hline
        \multicolumn{8}{c}{\textbf{Linguistic-Based Baselines}} \\
        \hline
        punctuation + capitalization & 74.2 & 67.7 & \textbf{98.5} & 80.2 & \textbf{96.5} & 46.4 & 62.7 \\
        bag-of-words & \textbf{79.1} & \textbf{76.4} & 88.0 & \textbf{81.8} & 83.4 & \textbf{69.1} & \textbf{75.6} \\
        fastText  & 64.2 & 63.5 & 69.4 & 66.3 & 65.2 & 59.0 & 61.9   \\
        \hline
        \multicolumn{8}{c}{\textbf{CNN/RNN-based}} \\
        \hline
        Char BiLSTM & \textbf{87.0}	& \textbf{80.9}	& \underline{\textbf{98.8}}	& \underline{\textbf{89.0}} & \underline{\textbf{98.1}}	& 73.5 & \textbf{84.0} \\
        
        Word BiLSTM (fastText) & 78.1 & 75.0 & 88.3 & 81.1 & 83.3 & 66.5 & 73.9 \\
        Universal Sentence Encoder+CNN & 85.6 & 80.5 & 95.8 & 87.5 & 89.4 & \textbf{80.7} & 82.5 \\
        ELMo & 84.6 & 79.6 & 95.6 & 86.9 & 93.6 & 72.1 & 81.4 \\
        \hline
        \multicolumn{8}{c}{\textbf{Transformer-based Encoders}} \\
        \hline
        BERT (uncased) & 77.4 & 72.8 & 92.1 & 81.4 & 87.1 & 60.6 & 71.4 \\
        BERT (cased) & 78.0 & 74.6 & 89.0 & 81.2 & 83.8 & 65.4 & 73.4\\
        DistilBERT (uncased) & 80.0 & 76.4 & 90.5 & 82.9 & 86.3 & 68.2 & \textbf{76.2}\\
        DistilBERT (cased) & 80.1 & 80.1 & 91.7 & 83.0 & 87.5 & 66.6 & 75.6 \\
        RoBERTa-base & 82.6 & 74.4 & 89.4 & 81.2 & 84.2 & 64.7 & 73.2\\
        DeBERTa-base & 87.2& 	83.7 & 	94.3 & 	88.7 & 	92.4 & 	79.0 & 85.2\\
        DeBERTa-large & \underline{\textbf{87.8}} & \underline{\textbf{85.0}} &	93.4 & \underline{\textbf{89.0}} &	91.6 &	\underline{\textbf{81.3}} &	\underline{\textbf{86.1}} \\  
         DeBERTaV3-large & 86.9 & 82.5& \textbf{95.7} &88.6& \textbf{94.0} & 76.9& 84.6\\
        \hline
        \multicolumn{8}{c}{\textbf{Transformer-based Decoders}} \\
        \hline
        GPT2  & 85.1 & 80.5 &  \textbf{95.1} & 87.2 & \textbf{92.9} & 73.5 & 82.1 \\
        XLNet  & \textbf{86.0} & \textbf{82.0} & 94.5 & \textbf{87.9} & 92.4 & \textbf{76.5} & \textbf{83.7} \\
        \bottomrule
    \end{tabular}
    \caption{Results of monolingual formality classification for English (GYAFC dataset). \textbf{Bold} numbers represents the best results in the category, \underline{\textbf{bold and underlined}} -- the best results for the metric.}
    \label{tab:gyafc}
\end{table*}
\endgroup




\section{Results}


\subsection{Experimental Setup}

Formality detection task could be cast as a binary classification task with classes \textsf{\small formal} and \textsf{\small informal}. Therefore, we report standard evaluation metrics for binary classification in experiments: Accuracy, Precision, Recall, and F1.

We report the results of three types of experiment setups to provide answers to three research questions mentioned in the introduction:
\begin{enumerate}
    \item \textit{Monolingual}: we fine-tune all mentioned in Section~\ref{sec:methods} type of models for monolingual English formality classification task and report Accuracy, Precision, Recall, and F1 scores; then, we use multilingual models to test them on four languages---English, Italian, Portuguese, and French---separately and report Accuracy for each language;
    \item \textit{Multilingual}: we fine-tune adapt some baselines and utilise mentioned multilingual pre-trained language models on all four languages and report total accuracy;
    \item \textit{Cross-lingual}: we fine-tune multilingual models on all languages except the target one (i.e. on English, Italian, Portuguese, but not French) and then perform zero-shot inference on the test set of that excluded from the training step language (i.e. French) reporting the Accuracy score.
\end{enumerate}

\subsection{Monolingual English Results}

\begingroup
\renewcommand{\arraystretch}{1.2}
\begin{table*}[h!]
    \centering
    \footnotesize
    \begin{tabular}{l|c|c|c|c|c}
        \toprule
        \textbf{Text Representation Model} & English & Italian & Portuguese & French & All \\
        \hline
        \multicolumn{6}{c}{\textbf{Linguistic-Based Baselines}} \\
        \hline
        punctuation + capitalization & 74.2 & 69.2 & 64.4 & 66.5 & 68.6 \\
        
        bag-of-words & \textbf{79.1} & \textbf{71.3} & \textbf{70.6} & \textbf{72.5} & -- \\
        fastText & 64.2 & 56.0 & 54.3 & 58.6 & --  \\
        \hline
        \multicolumn{6}{c}{\textbf{CNN/RNN-based}} \\
        \hline

        Char BiLSTM & \textbf{87.0} & \underline{\textbf{79.1}} & \textbf{75.9}	 & \underline{\textbf{81.3}} & \underline{\textbf{82.7}}  \\        
        Word BiLSTM (fastText) & 78.1 & 68.7 & 68.9 & 69.2 & 70.2 \\
        Universal Sentence Encoder+CNN & 85.4 & 76.7 & 75.3 & 80.7 & 80.0 \\
        \hline
        \multicolumn{6}{c}{\textbf{Transformer-based Encoders}} \\
        \hline

        mBERT (uncased)           & 70.9 & 72.3 & 72.3 & 73.1 & 74.7 \\        
        mBERT (cased)             & 83.0 & \textbf{77.8} & \underline{\textbf{77.3}} & \textbf{79.9} & \textbf{79.9} \\    
        mDistilBERT (cased)            & 86.6	&76.8	&75.9&	79.1&	79.4 \\  

         mDeBERTaV3-base	& \underline{\textbf{87.3}}	&76.6	&75.8&	78.9&	\textbf{79.9} \\    \hline
       
        \multicolumn{6}{c}{\textbf{Transformer-based Decoders}} \\
        \hline
        XLM-R  & 85.2 & \textbf{76.9} & \textbf{76.2} & \textbf{79.5} & \textbf{79.4} \\ 
        mT5-base & 83.4 &	72.9 &	70.3 &	72.4 & 78.2 \\ 
        mBART-large & \textbf{86.9}	& \textbf{76.9} &	75.9	&79.3&	79.0 \\
         
        \bottomrule
    \end{tabular}
    \caption{Accuracy results of both monolingual and multilingual formality classification for English, Italian, Portuguese, and French (X-FORMAL dataset).  Here ``All'' denotes that the model was trained and tested on all presented languages. \textbf{Bold} numbers represents the best results in the category, \underline{\textbf{bold and underlined}} -- the best results for the metric.}
    \label{tab:xformal}
\end{table*}
\endgroup

\begingroup
\renewcommand{\arraystretch}{1.2}
\begin{table*}[h!]
    \centering
    \footnotesize
    \begin{tabular}{l|c|c|c|c}
        \toprule
        \textbf{Train / Test} & English & Italian & Portuguese & French  \\


        \hline
        \multicolumn{5}{c}{\textbf{Universal Sentence Encoder}} \\ \hline
        Monolingual & 85.4 &  \textbf{76.7} & \textbf{75.3} & \textbf{80.7}  \\
        \hline
        All but English & 77.5 & - & - & - \\
        All but Italian  & - & 72.6 & - & -  \\ 
        All but Portugese & - & - & 70.5 & - \\
        All but French & - & - & - & 72.6 \\ 
        \hline
        All & \textbf{85.9} & 76.5 & 75.0 & 79.0 \\ 
        
        \hline
        \multicolumn{5}{c}{\textbf{mBERT (cased)}} \\ \hline
        Monolingual & \textbf{83.0} & \textbf{77.8} & \textbf{77.3} & \textbf{79.9} \\
        \hline
        All but English & 79.9 & - & - & - \\
        All but Italian & - & 73.0 & - & -  \\ 
        All but Portugese & - & - & 71.6 & - \\
        All but French & - & - & - & 71.6 \\ 
        \hline
        All  & 80.2 & 73.1 & 72.2 & 75.0 \\
        
        \hline
        \multicolumn{5}{c}{\textbf{Char BiLSTM}} \\ \hline
        Monolingual  & \underline{\textbf{87.0}} & \underline{\textbf{79.1}} & \underline{\textbf{75.9}} & \underline{\textbf{81.3}} \\
        \hline
        All but English & 74.9 & - & - & - \\
        All but Italian & - & 74.1 & - & -  \\ 
        All but Portugese & - & - & 71.9 & - \\
        All but French & - & - & - & \underline{77.4} \\ 
        \hline
        All  & 83.1 & 75.2 & 74.2 & 78.0 \\

        \hline
        \multicolumn{5}{c}{\textbf{mDistilBERT (cased) }} \\ \hline
        Monolingual  & \textbf{86.6} & \textbf{76.8} & \textbf{75.9} &	\textbf{79.4} \\
        \hline
        All but English & \underline{83.6} & - & - & - \\
        All but Italian & - & \underline{75.1} & - & -  \\ 
        All but Portugese & - & - & \underline{73.8} & - \\
        All but French & - & - & - & 77.1 \\ 
        \hline
        All  & 85.9	& \textbf{76.8}	& \textbf{75.9} &	79.1 \\

        \bottomrule
    \end{tabular}
    \caption{Accuracy results of cross-language transfer study on formality classification. \textbf{Bold} numbers represents the best results for the model type, \underline{underlined} -- the best results for cross-lingual transfer to the language, \underline{\textbf{bold and underlined}} -- the best results for the language.}
    \label{tab:cross_lang}
\end{table*}
\endgroup

Firstly, we present monolingual formality classification results on English GYAFC corpus. Results of the experiments with the various models described in Section~\ref{sec:methods}  are presented in Table~\ref{tab:gyafc}. 

\paragraph{Ranking of the models} Firstly, we can observe already quite high results for the simple baseline models. The classification approach based on punctuation and capitalization presence features achieves one of the highest results for the formal class Recall score$=98.5$, however failed to distinguish informal class so well (Recall$=46.4$). Bag-of-words approach reaches F1 scores for both classes on the level with Transformer-based models ($81.8$ and $75.6$ respectfully).

A significant number of Convolution-based Neural Networks exhibit superior performance in comparison to the baseline models, with certain models showcasing a notable gap in performance. Particularly, the Char BiLSTM model surpasses all other models within this category and achieves remarkably high scores across all evaluation metrics. This model excels in terms of formal class Recall and F1 scores and informal class Precision ($98.8$, $89.0$, and $98.1$ respectfully).

Among the category of classification models based on Transformers, a substantial proportion of these models exhibit notable performance, with encoder-based architectures demonstrating a slight superiority over decoder-based ones. Although certain BERT models do not surpass certain baseline models, the succeeding next generation of BERT-based models yield high performance across all evaluation metrics. Notably, within the category of Transformer-based pre-trained language models, DeBERTa attains the highest performance results among all compared models in terms of total Accuracy$=87.8$ and F1 scores for both classes ($89.0$ for formal and $86.1$ for informal).


This brings us to the answer of the question \textbf{Q1}: Deep pre-trained models like DeBERTa yield top performance for monolingual English formality classification task. At the same time, Char BiLSTM model yield as well superior results for some metrics even outperforming DeBERTa.

\paragraph{Impact of case-sensitivity} Within the several type of models we can observe that capitalization sensitivity is quite important for formality detection task. As such, for linguistic-based baseline, these features prove highly effective in attaining high scores, particularly for formal class. We can also compare cased and uncased versions for BERT and DistilBERT models. Although cased models demonstrate a superiority in terms of Accuracy scores ($78.0$ vs $77.4$ and $80.1$ vs $80.0$), the results of other metrics do not establish a clear and definitive winner.

\subsection{Monolingual and Multilingual Results for Four Languages}
In this section, we report results on the X-FORMAL dataset~\cite{briakou-etal-2021-ola}. Results of the experiments with the various models described in Section~\ref{sec:methods}  presented in Table~\ref{tab:xformal}. 

\paragraph{Monolingual results} Firstly, we conducted experiments exploring multilingual models for monolingual classification for all languages separately -- English, Italian, Portuguese, and French. As one may observe, similarly to English results, the model based on a bidirectional LSTM model with character embeddings yields the best results for all languages. Some multilingual transformer-based models such as XLM-R and mBERT also achieve good enough results but are lower than Char BiLSTM. Except Portuguese language, where mBART (cased) model has the highest accuracy. 
%

\paragraph{Multilingual results} We report the results of fine-tuned multilingual language models on all provided languages in ``All'' column in Table~\ref{tab:xformal} and inference of these models on each language separately in Table~\ref{tab:cross_lang}. For all best models across different categories, we can  observer a slight drop of the accuracy for all languages in comparison to monolingual results. For instance, for the best performing model Char BiLSTM, the ``All'' Accuracy$=82.7$ is less then monolingual setups: English ($83.1$ vs $87.0$), Italian ($75.2$ vs $79.1$), Portuguese ($74.2$ vs $75.9$), French ($78.0$ vs $81.3$). However, these drops in the Accuracy scores is slight and the scores outperform the monolingual baselines and some Transformer-based models significantly.

As a result, the simultaneous fine-tuning of multilingual formality detection models does not cause a significant drop of the performance across languages in comparison of the best monolingual results. The high results of multilingual Char BiLSTM model provides a positive answer to the question \textbf{Q2}.


\subsection{Cross-lingual Formality Transfer Results} 

After multilingual experiments, we conducted cross-lingual ones trying to answer the research question \textbf{Q3}. The results of the experiments are presented in Table~\ref{tab:cross_lang}. The main conclusion that can be made from the obtained results is that cross-lingual formality detection is possible but, unfortunately, the same as for multilingual results, with a drop in the performance across languages. For all reported models, we can observe the drop of Accuracy scores in $3-5\%$.

For the best performing models from previously discussed monolingual and multilingual results---Char BiLSTM---we can observe a significant drop in the performance in comparison to its best results. However, mDistilBERT demonstrates more stable performance to unseen languages in the training set. This model has the best cross-lingual formality transfer capability with achieving cross-lingual English Accuracy$=83.6$ (vs only $74.9$ from Char BiLSTM), Italian Accuracy$=75.1$ (vs $74.1$ from Char BiLSTM), Portuguese Accuracy$=73.8$ (vs $71.9$ from Char BiLSTM), and only for French Accuracy$=77.1$, Char BiLSTM model shows slightly better performance with Accuracy$=77.4$.

Despite the loss in accuracy compared to the best monolingual results, the illustrated results of cross-lingual experiments again provide a positive answer to the stated question \textbf{Q3}. Still, the cross-lingual tests of the best performing models overcomes the monolingual baselines. This implies the possibility to the cross-lingual formality transfer usage to perform classification on the unseen language with satisfactory accuracy.

\section{Discussions}
As all the above experiments results showed that none of the models achieved Accuracy and F1 scores higher $90.0$, we analyzed misclassifications. In Appendix~\ref{sec:app_misclassified} in Table \ref{tab:errors_examples}, we present several examples of such models mistakes. We noticed that the misclassification of formal sentences into informal appeared less often than informal into formal which confirms with high Recall scores for formal class and significantly lower scores for informal one in Table~\ref{tab:gyafc}. For example, for the DeBERTa-large model, the rate of misclassification of formal sentences into informal is only $6.6\%$, while misclassification of informal sentences into formal -- $18.7\%$. Some of the mistakes are connected with the unobvious labels of the original data. 

For example, the Char BiLSTM model trained for the English language misclassified sentence \textit{1 WOULD WORK FOR ME BUT BOTH WOULD BE EVEN BETTER} into formal class. Indeed, the whole structure of the sentence and the usage of word \textit{would} make the text looks like a formal one. We suppose that this text was marked as informal because it is fully written in the upper register. 

On the other hand, there are many sentences with formal labels without an obvious reason for that. Texts like \textit{Ignore it when people start rumors.}, \textit{I do not want her to die.} does not look like to be written in a formal style. On the contrary, the usage of the phrase \textit{Ignore it} seems to be quite informal. 


Also, if we look at misclassification examples of mDistillBERT models, we can see examples of obvious violations of formal style. For example, we can observe sentences that are grammatically correct, but the content is toxic (\textit{Are you serious or just that ignorant?}) or refers to some informal ways of entertainment (\textit{After watching that, I had to consume alcohol!}). That might be that the general topic of these sentences is more closer to the topics usually discussed informally that confuses the model. In addition, we draw attention to the sample which is mostly formal, however, contains informal insertion: \textit{I’m grateful, I now comprehend. Significantly, er, electrical.}

Such mistakes can be connected with the process of the creation of the GYAFC and XFORMAL datasets. The train part consists of informal texts and their formal paraphrases with Amazon Mechanical Turk workers. However, the tune part contains paraphrases from formal into informal styles and vice versa. The annotation process can contain some inaccuracies that may be resulting in fuzzy logic of labels assignment.

In addition, another interesting observation might be that for some Transformer-based models their multilingual versions yields higher accuracy than monolingual ones. Thus, for DistilBERT, the bets English monolingual Accuracy is $80.1$, while its multilingual version achieves $86.6$ score on English test set. The same observation can be applied for BERT model as well.

In the end, we can observe quit high results from Char BiLSTM model which outperform in some cases Transformer-based models. One of the explanations might be: the usage of slang or unusually modified words in informal style that can be precisely tokenized and embedded with Transformer-based encoders, however, can be learned with character-level words' split.

\section{Conclusion}

In this paper, we presented the first computational study on text categorization models that detect text formality. We based our experiments on two large-scale multilingual datasets---GYAFC and X-FORMAL---and tested a vast amount of baselines and state-of-the-art neural models.

The best English monolingual results are achieved by Transformer-based model---DeBERTa-large. However, other obtained results show the superiority of models based on character representation, such as Char BiLSTM models, over models based on word and BPE representations, including even large pre-trained transformer models. Notably for both monolingual and multilingual formality detection for all examined languages, Char BiLSTM model illustrates the best accuracy.  

Our experiments also show that multiple models demonstrate abilities of cross-lingual transfer. While Char BiLSTM showed the best performance in monolingual and multilingual setups, it had a significant drop in the performance while trying to transfer formality knowledge to another language. In this scenario, mDistilBERT model demonstrated the best stability to new languages.



All code and data allowing reproduce our experiments are available online.\footnote{\href{https://github.com/s-nlp/formality}{https://github.com/s-nlp/formality}} We release for a public usage the best Transformer-based monolingual\footnote{\href{https://huggingface.co/s-nlp/deberta-large-formality-ranker}{https://huggingface.co/s-nlp/deberta-large-formality-ranker}}, multilingual\footnote{\href{https://huggingface.co/s-nlp/mdeberta-base-formality-ranker}{https://huggingface.co/s-nlp/mdeberta-base-formality-ranker}}, and cross-lingual\footnote{\href{https://huggingface.co/s-nlp/mdistilbert-base-formality-ranker}{https://huggingface.co/s-nlp/mdistilbert-base-formality-ranker}} models.

\section*{Acknowledgments}

We thank Andrey Likhachev and Ivan Trifinov for preparing the first version of the experiments presented in this paper. This research was partially supported by a joint MTS-Skoltech lab on AI and the European Union's Horizon 2020 research and innovation program under the Marie Skodowska-Curie grant agreement No 860621, and the Galician Ministry of Culture, Education, Professional Training, and University and the European Regional Development Fund (ERDF/FEDER program) under grant ED431G2019/04. Additionaly, we are grateful Social Research Computing Group, TUM for the possibility to finalize this research.


\section{Ethical Statement}


We hope that models' research in formality classification and style transfer tasks might help to develop more sophisticated approaches for language and style studying programs. For instance, such an automated helper can detect incorrect style used for a text exercise, explain a style misusage, and recommend a correct paraphrase. This may be useful for language learners who do not realize nuances of language at the level of native speakers preventing their deeper integration in a given society. 

Furthermore, the availability of formality data in four languages provides a solid foundation and we have shown that the cross-lingual formality detection is possible. We anticipate that research in the field of formality detection foster development of  similar datasets in other languages as well.

Last but not least, our approach and experiments are based on large pre-trained language models, which may be prone to biases reflected in their training data. In case of real world deployments this issue shall be taken into account. 

\bibliographystyle{acl_natbib}
\bibliography{ranlp2023}

\begin{thebibliography}{31}
\expandafter\ifx\csname natexlab\endcsname\relax\def\natexlab#1{#1}\fi

\bibitem[{Bojanowski et~al.(2017)Bojanowski, Grave, Joulin, and
  Mikolov}]{bojanowski-etal-2017-enriching}
Piotr Bojanowski, Edouard Grave, Armand Joulin, and Tomas Mikolov. 2017.
\newblock \href {https://aclanthology.org/Q17-1010} {Enriching word vectors
  with subword information}.
\newblock \emph{Transactions of the Association for Computational Linguistics},
  5:135--146.

\bibitem[{Briakou et~al.(2021)Briakou, Lu, Zhang, and
  Tetreault}]{briakou-etal-2021-ola}
Eleftheria Briakou, Di~Lu, Ke~Zhang, and Joel Tetreault. 2021.
\newblock \href {https://doi.org/10.18653/v1/2021.naacl-main.256} {Ol{\'a},
  bonjour, salve! {XFORMAL}: A benchmark for multilingual formality style
  transfer}.
\newblock In \emph{Proceedings of the 2021 Conference of the North American
  Chapter of the Association for Computational Linguistics: Human Language
  Technologies}, pages 3199--3216, Online. Association for Computational
  Linguistics.

\bibitem[{Cer et~al.(2018)Cer, Yang, Kong, Hua, Limtiaco, St.~John, Constant,
  Guajardo-Cespedes, Yuan, Tar, Strope, and Kurzweil}]{cer-etal-2018-universal}
Daniel Cer, Yinfei Yang, Sheng-yi Kong, Nan Hua, Nicole Limtiaco, Rhomni
  St.~John, Noah Constant, Mario Guajardo-Cespedes, Steve Yuan, Chris Tar,
  Brian Strope, and Ray Kurzweil. 2018.
\newblock \href {https://doi.org/10.18653/v1/D18-2029} {Universal sentence
  encoder for {E}nglish}.
\newblock In \emph{Proceedings of the 2018 Conference on Empirical Methods in
  Natural Language Processing: System Demonstrations}, pages 169--174,
  Brussels, Belgium. Association for Computational Linguistics.

\bibitem[{Chidambaram et~al.(2019)Chidambaram, Yang, Cer, Yuan, Sung, Strope,
  and Kurzweil}]{chidambaram2019learning}
Muthuraman Chidambaram, Yinfei Yang, Daniel Cer, Steve Yuan, Yun-Hsuan Sung,
  Brian Strope, and Ray Kurzweil. 2019.
\newblock \href {http://arxiv.org/abs/1810.12836} {Learning cross-lingual
  sentence representations via a multi-task dual-encoder model}.

\bibitem[{Conneau et~al.(2020)Conneau, Khandelwal, Goyal, Chaudhary, Wenzek,
  Guzm{\'a}n, Grave, Ott, Zettlemoyer, and
  Stoyanov}]{conneau-etal-2020-unsupervised}
Alexis Conneau, Kartikay Khandelwal, Naman Goyal, Vishrav Chaudhary, Guillaume
  Wenzek, Francisco Guzm{\'a}n, Edouard Grave, Myle Ott, Luke Zettlemoyer, and
  Veselin Stoyanov. 2020.
\newblock \href {https://doi.org/10.18653/v1/2020.acl-main.747} {Unsupervised
  cross-lingual representation learning at scale}.
\newblock In \emph{Proceedings of the 58th Annual Meeting of the Association
  for Computational Linguistics}, pages 8440--8451, Online. Association for
  Computational Linguistics.

\bibitem[{Devlin et~al.(2018)Devlin, Chang, Lee, and
  Toutanova}]{devlin2018bert}
Jacob Devlin, Ming-Wei Chang, Kenton Lee, and Kristina Toutanova. 2018.
\newblock {BERT}: Pre-training of deep bidirectional transformers for language
  understanding.
\newblock \emph{arXiv preprint arXiv:1810.04805}.

\bibitem[{Devlin et~al.(2019)Devlin, Chang, Lee, and
  Toutanova}]{devlin2019bert}
Jacob Devlin, Ming-Wei Chang, Kenton Lee, and Kristina Toutanova. 2019.
\newblock {BERT}: Pre-training of deep bidirectional transformers for language
  understanding.
\newblock In \emph{NAACL-HLT (1)}.

\bibitem[{Graves and Schmidhuber(2005)}]{graves2005framewise}
Alex Graves and J{\"u}rgen Schmidhuber. 2005.
\newblock Framewise phoneme classification with bidirectional {LSTM} and other
  neural network architectures.
\newblock \emph{Neural networks}, 18(5-6):602--610.

\bibitem[{Hameed and Garcia{-}Zapirain(2020)}]{DBLP:journals/access/HameedG20}
Zabit Hameed and Begonya Garcia{-}Zapirain. 2020.
\newblock \href {https://doi.org/10.1109/ACCESS.2020.2988550} {Sentiment
  classification using a single-layered bilstm model}.
\newblock \emph{{IEEE} Access}, 8:73992--74001.

\bibitem[{Hayes et~al.(1988)Hayes, Knecht, and Cellio}]{hayes1988news}
Philip~J Hayes, Laura~E Knecht, and Monica~J Cellio. 1988.
\newblock A news story categorization system.
\newblock In \emph{Second Conference on Applied Natural Language Processing},
  pages 9--17.

\bibitem[{He et~al.(2021)He, Liu, Gao, and Chen}]{DBLP:conf/iclr/HeLGC21}
Pengcheng He, Xiaodong Liu, Jianfeng Gao, and Weizhu Chen. 2021.
\newblock \href {https://openreview.net/forum?id=XPZIaotutsD} {{DeBERTa}:
  decoding-enhanced bert with disentangled attention}.
\newblock In \emph{9th International Conference on Learning Representations,
  {ICLR} 2021, Virtual Event, Austria, May 3-7, 2021}. OpenReview.net.

\bibitem[{Isnain et~al.(2020)Isnain, Sihabuddin, and
  Suyanto}]{isnain2020bidirectional}
Auliya~Rahman Isnain, Agus Sihabuddin, and Yohanes Suyanto. 2020.
\newblock Bidirectional long short term memory method and word2vec extraction
  approach for hate speech detection.
\newblock \emph{IJCCS (Indonesian Journal of Computing and Cybernetics
  Systems)}, 14(2):169--178.

\bibitem[{Joos(1976)}]{joos1976five}
Martin Joos. 1976.
\newblock \emph{Five Clocks Times}.
\newblock Washington DC: Georgetown University Press.

\bibitem[{Lai et~al.(2022)Lai, Mao, Toral, and Nissim}]{lai-etal-2022-human}
Huiyuan Lai, Jiali Mao, Antonio Toral, and Malvina Nissim. 2022.
\newblock \href {https://doi.org/10.18653/v1/2022.humeval-1.9} {Human judgement
  as a compass to navigate automatic metrics for formality transfer}.
\newblock In \emph{Proceedings of the 2nd Workshop on Human Evaluation of NLP
  Systems (HumEval)}, pages 102--115, Dublin, Ireland. Association for
  Computational Linguistics.

\bibitem[{Lewis(1991)}]{lewis1991evaluating}
David~D Lewis. 1991.
\newblock Evaluating text categorization i.
\newblock In \emph{Speech and Natural Language: Proceedings of a Workshop Held
  at Pacific Grove, California, February 19-22, 1991}.

\bibitem[{Li et~al.(2022)Li, Wang, Balducci, Hu, Gordon, Marinova, and
  Shang}]{DBLP:conf/iri/LiWBHGMS22}
Can Li, Wenbo Wang, Bitty Balducci, Lingshu Hu, Matthew Gordon, Detelina
  Marinova, and Yi~Shang. 2022.
\newblock \href {https://doi.org/10.1109/IRI54793.2022.00014} {Deep formality:
  Sentence formality prediction with {D}eep {L}earning}.
\newblock In \emph{23rd {IEEE} International Conference on Information Reuse
  and Integration for Data Science, {IRI} 2022, San Diego, CA, USA, August
  9-11, 2022}, pages 1--5. {IEEE}.

\bibitem[{Liu et~al.(2019)Liu, Ott, Goyal, Du, Joshi, Chen, Levy, Lewis,
  Zettlemoyer, and Stoyanov}]{DBLP:journals/corr/abs-1907-11692}
Yinhan Liu, Myle Ott, Naman Goyal, Jingfei Du, Mandar Joshi, Danqi Chen, Omer
  Levy, Mike Lewis, Luke Zettlemoyer, and Veselin Stoyanov. 2019.
\newblock \href {http://arxiv.org/abs/1907.11692} {Ro{BERT}a: {A} robustly
  optimized {BERT} pretraining approach}.
\newblock \emph{CoRR}, abs/1907.11692.

\bibitem[{Pavlick and Tetreault(2016)}]{pavlick2016empirical}
Ellie Pavlick and Joel Tetreault. 2016.
\newblock An empirical analysis of formality in online communication.
\newblock \emph{Transactions of the Association for Computational Linguistics},
  4:61--74.

\bibitem[{Peters et~al.(2018)Peters, Neumann, Iyyer, Gardner, Clark, Lee, and
  Zettlemoyer}]{peters-etal-2018-deep}
Matthew~E. Peters, Mark Neumann, Mohit Iyyer, Matt Gardner, Christopher Clark,
  Kenton Lee, and Luke Zettlemoyer. 2018.
\newblock \href {https://doi.org/10.18653/v1/N18-1202} {Deep contextualized
  word representations}.
\newblock In \emph{Proceedings of the 2018 Conference of the North {A}merican
  Chapter of the Association for Computational Linguistics: Human Language
  Technologies, Volume 1 (Long Papers)}, pages 2227--2237, New Orleans,
  Louisiana. Association for Computational Linguistics.

\bibitem[{Radford et~al.(2019)Radford, Wu, Child, Luan, Amodei, Sutskever
  et~al.}]{radford2019language}
Alec Radford, Jeffrey Wu, Rewon Child, David Luan, Dario Amodei, Ilya
  Sutskever, et~al. 2019.
\newblock Language models are unsupervised multitask learners.
\newblock \emph{OpenAI blog}, 1(8):9.

\bibitem[{Rao and Tetreault(2018)}]{rao-tetreault-2018-dear}
Sudha Rao and Joel Tetreault. 2018.
\newblock \href {https://doi.org/10.18653/v1/N18-1012} {Dear sir or madam, may
  {I} introduce the {GYAFC} dataset: Corpus, benchmarks and metrics for
  formality style transfer}.
\newblock In \emph{Proceedings of the 2018 Conference of the North {A}merican
  Chapter of the Association for Computational Linguistics: Human Language
  Technologies, Volume 1 (Long Papers)}, pages 129--140, New Orleans,
  Louisiana. Association for Computational Linguistics.

\bibitem[{Sanh et~al.(2019)Sanh, Debut, Chaumond, and
  Wolf}]{DBLP:journals/corr/abs-1910-01108}
Victor Sanh, Lysandre Debut, Julien Chaumond, and Thomas Wolf. 2019.
\newblock \href {http://arxiv.org/abs/1910.01108} {{DistilBERT}, a distilled
  version of {BERT:} smaller, faster, cheaper and lighter}.
\newblock \emph{CoRR}, abs/1910.01108.

\bibitem[{Sebastiani(2002)}]{sebastiani2002machine}
Fabrizio Sebastiani. 2002.
\newblock Machine learning in automated text categorization.
\newblock \emph{ACM computing surveys (CSUR)}, 34(1):1--47.

\bibitem[{Sun et~al.(2019)Sun, Qiu, Xu, and Huang}]{sun2019fine}
Chi Sun, Xipeng Qiu, Yige Xu, and Xuanjing Huang. 2019.
\newblock How to fine-tune {BERT} for text classification?
\newblock In \emph{China National Conference on Chinese Computational
  Linguistics}, pages 194--206. Springer.

\bibitem[{Tang et~al.(2020)Tang, Tran, Li, Chen, Goyal, Chaudhary, Gu, and
  Fan}]{tang2020multilingual}
Yuqing Tang, Chau Tran, Xian Li, Peng-Jen Chen, Naman Goyal, Vishrav Chaudhary,
  Jiatao Gu, and Angela Fan. 2020.
\newblock \href {http://arxiv.org/abs/2008.00401} {Multilingual translation
  with extensible multilingual pretraining and finetuning}.

\bibitem[{Vaswani et~al.(2017)Vaswani, Shazeer, Parmar, Uszkoreit, Jones,
  Gomez, Kaiser, and Polosukhin}]{DBLP:conf/nips/VaswaniSPUJGKP17}
Ashish Vaswani, Noam Shazeer, Niki Parmar, Jakob Uszkoreit, Llion Jones,
  Aidan~N. Gomez, Lukasz Kaiser, and Illia Polosukhin. 2017.
\newblock \href
  {https://proceedings.neurips.cc/paper/2017/hash/3f5ee243547dee91fbd053c1c4a845aa-Abstract.html}
  {Attention is all you need}.
\newblock In \emph{Advances in Neural Information Processing Systems 30: Annual
  Conference on Neural Information Processing Systems 2017, December 4-9, 2017,
  Long Beach, CA, {USA}}, pages 5998--6008.

\bibitem[{Wiedemann et~al.(2018)Wiedemann, Ruppert, Jindal, and
  Biemann}]{DBLP:journals/corr/abs-1811-02906}
Gregor Wiedemann, Eugen Ruppert, Raghav Jindal, and Chris Biemann. 2018.
\newblock \href {http://arxiv.org/abs/1811.02906} {Transfer learning from {LDA}
  to bilstm-cnn for offensive language detection in twitter}.
\newblock \emph{CoRR}, abs/1811.02906.

\bibitem[{Xue et~al.(2021)Xue, Constant, Roberts, Kale, Al{-}Rfou, Siddhant,
  Barua, and Raffel}]{xue2021mt5}
Linting Xue, Noah Constant, Adam Roberts, Mihir Kale, Rami Al{-}Rfou, Aditya
  Siddhant, Aditya Barua, and Colin Raffel. 2021.
\newblock \href {https://doi.org/10.18653/v1/2021.naacl-main.41} {mt5: {A}
  massively multilingual pre-trained text-to-text transformer}.
\newblock In \emph{Proceedings of the 2021 Conference of the North American
  Chapter of the Association for Computational Linguistics: Human Language
  Technologies, {NAACL-HLT} 2021, Online, June 6-11, 2021}, pages 483--498.
  Association for Computational Linguistics.

\bibitem[{Yang et~al.(2019{\natexlab{a}})Yang, Cer, Ahmad, Guo, Law, Constant,
  {\'{A}}brego, Yuan, Tar, Sung, Strope, and
  Kurzweil}]{DBLP:journals/corr/abs-1907-04307}
Yinfei Yang, Daniel Cer, Amin Ahmad, Mandy Guo, Jax Law, Noah Constant,
  Gustavo~Hern{\'{a}}ndez {\'{A}}brego, Steve Yuan, Chris Tar, Yun{-}Hsuan
  Sung, Brian Strope, and Ray Kurzweil. 2019{\natexlab{a}}.
\newblock \href {http://arxiv.org/abs/1907.04307} {Multilingual universal
  sentence encoder for semantic retrieval}.
\newblock \emph{CoRR}, abs/1907.04307.

\bibitem[{Yang et~al.(2019{\natexlab{b}})Yang, Dai, Yang, Carbonell,
  Salakhutdinov, and Le}]{DBLP:conf/nips/YangDYCSL19}
Zhilin Yang, Zihang Dai, Yiming Yang, Jaime~G. Carbonell, Ruslan Salakhutdinov,
  and Quoc~V. Le. 2019{\natexlab{b}}.
\newblock \href
  {https://proceedings.neurips.cc/paper/2019/hash/dc6a7e655d7e5840e66733e9ee67cc69-Abstract.html}
  {{XLN}et: Generalized autoregressive pretraining for language understanding}.
\newblock In \emph{Advances in Neural Information Processing Systems 32: Annual
  Conference on Neural Information Processing Systems 2019, NeurIPS 2019,
  December 8-14, 2019, Vancouver, BC, Canada}, pages 5754--5764.

\bibitem[{Zhang et~al.(2015)Zhang, Zhao, and LeCun}]{DBLP:conf/nips/ZhangZL15}
Xiang Zhang, Junbo~Jake Zhao, and Yann LeCun. 2015.
\newblock \href
  {https://proceedings.neurips.cc/paper/2015/hash/250cf8b51c773f3f8dc8b4be867a9a02-Abstract.html}
  {Character-level {C}onvolutional {N}etworks for text classification}.
\newblock In \emph{Advances in Neural Information Processing Systems 28: Annual
  Conference on Neural Information Processing Systems 2015, December 7-12,
  2015, Montreal, Quebec, Canada}, pages 649--657.

\end{thebibliography}

\onecolumn

\appendix
\section{Classification Error Analysis}
\label{sec:app_misclassified}
Here, we provide the misclassification results for one the best performing models for English monolingual classification--Char BiLSTM, the best Transformer-based monolingual model---DeBERTa-large---and the best model with cross-lingual formality transfer capabilities--mDistilBERT.

\begingroup
\renewcommand{\arraystretch}{1.2}
\begin{table*}[h!]
    \centering
    \begin{tabular}{p{10.5cm}|c|c}
        \toprule
        \textbf{Sentence} & \textbf{Original Label} & \textbf{Predicted Label} \\
        \hline
        \multicolumn{3}{c}{\textbf{Char BiLSTM}} \\
        \hline
        That has 2 b the worst hiding spot ever. & Formal & Informal \\
        I would not be mad at you forever. & Formal & Informal \\
        No, he doesn't even know her. They met online. & Formal & Informal \\
        I tune in to lotsa music. & Formal & Informal \\
        I hate wearin flats, i aint gunna wear em for a guy. & Formal & Informal \\
        He is nice, but I have to question his thinking skills. & Informal & Formal \\
        Perhaps they were concerned that if you knew, you would be angry.. & Informal & Formal \\
        having fun is most important. & Informal & Formal \\
        Hold on a moment and let me think. & Informal & Formal \\
        Americans this is the aircraft carrier U.S.S. Lincoln, the second largest ship in the United States Atlantic fleet. & Informal & Formal \\
        \hline

        \multicolumn{3}{c}{\textbf{DeBERTa}} \\
        \hline
        It appears that they are going to turn it into a television series. & Formal & Informal \\
        Any film in which Johnny Depp appears. & Formal & Informal \\
        The song was Played on the Radio by Green Day. & Formal & Informal \\
        You need to sign another paper everyday with eachother. & Formal & Informal \\
        Not love, but who knows? & Formal & Informal \\
        and for everyone's information it was NOT geeky!!!! & Informal & Formal \\
        Someone watches him every move now! & Informal & Formal \\
        U come and go , come and go. & Informal & Formal \\
        But yes, this show is addicting! & Informal & Formal \\
        Run like hell and never look back. & Informal & Formal \\
        \hline

        \hline
        \multicolumn{3}{c}{\textbf{mDistilBERT}} \\
        \hline
        Don't spend your money on frivolous things. & Formal & Informal \\
        Are you serious or just that ignorant? & Formal & Informal \\
        I'm grateful, I now comprehend. Significantly, er, electrical. & Formal & Informal \\
        After watching that, I had to consume alcohol! & Formal & Informal \\
        What can I do when I see her being so upset? & Formal & Informal \\
        I want my budz to give me this gift like it's Christmas. & Informal & Formal \\
        can't remember the site, but if u need more miles lemme know, I have a lot & Informal & Formal \\
        i would stop calling and see if he misses you and calls you! & Informal & Formal \\
        You can look but You cant find. & Informal & Formal \\
        You aren't asking anything really. & Informal & Formal \\

        \bottomrule
    \end{tabular}
    \caption{Examples of top-models' errors on GYAFC dataset.}
    \label{tab:errors_examples}
\end{table*}
\endgroup

\end{document}